\documentclass[conference]{IEEEtran}
\IEEEoverridecommandlockouts
\usepackage{cite}
\usepackage{amsmath,amssymb,amsfonts}
\usepackage{algorithmic}
\usepackage{graphicx}
\usepackage{textcomp}
\usepackage{xcolor}
\usepackage{array}

\usepackage{enumerate}
\usepackage{url}
\usepackage{multirow}
\usepackage{booktabs}

\def\BibTeX{{\rm B\kern-.05em{\sc i\kern-.025em b}\kern-.08em
    t\kern-.1667em\lower.7ex\hbox{E}\kern-.125emX}}

\usepackage{fancyhdr}
\begin{document}

\title{Attending to Emotional Narratives
\thanks{We acknowledge funding from the A*STAR Human-Centric Artificial Intelligence Programme (SERC SSF Project No. A1718g0048), a Stanford IRiSS Computational Social Science Fellowship to DCO, and NIH Grant 1R01MH112560-01 to JZ. Correspondence to DCO at dco@comp.nus.edu.sg}
} 

\author{

\IEEEauthorblockN{Zhengxuan~Wu\textsuperscript{1}, Xiyu~Zhang\textsuperscript{2}, Tan~Zhi-Xuan\textsuperscript{3}, Jamil~Zaki\textsuperscript{4}, Desmond~C.~Ong\textsuperscript{3,5}}
\IEEEauthorblockA{\textit{\textsuperscript{1}Department of Management Science and Engineering, Stanford University, Stanford CA, USA} \\
\textit{\textsuperscript{2}Department of Computer Science, Stanford University, Stanford CA, USA} \\
\textit{\textsuperscript{3}A*STAR Artificial Intelligence Initiative, Agency for Science, Technology and Research, Singapore} \\
\textit{\textsuperscript{4}Department of Psychology, Stanford University, Stanford CA, USA} \\
\textit{\textsuperscript{5}Department of Information Systems and Analytics, National University of Singapore} \\
\{wuzhengx, sherinez\}@stanford.edu,  xuan@mit.edu, jzaki@stanford.edu,  
dco@comp.nus.edu.sg}

}

\maketitle
\thispagestyle{fancy}

\begin{abstract}

Attention mechanisms in deep neural networks have achieved excellent performance on sequence-prediction tasks. 
Here, we show that these recently-proposed attention-based mechanisms---in particular, the \emph{Transformer} with its parallelizable self-attention layers, and the \emph{Memory Fusion Network} with attention across modalities and time---also generalize well to multimodal time-series emotion recognition. 
Using a recently-introduced dataset of emotional autobiographical narratives, 
we adapt and apply these two attention mechanisms to predict emotional valence over time.
Our models perform extremely well, in some cases reaching a performance comparable with human raters. We end with a discussion of the implications of attention mechanisms to affective computing. 


\end{abstract}


\section{Introduction}


Imagine meeting a group of friends for a conversation over drinks: In such social situations, we often share anecdotes and stories from our lives. Being able to attend to these stories and understand our friends' emotions is a fundamentally human skill that people possess and effortlessly employ. Such social understanding enables us to reason about the feelings of those around us, empathize with them, and build strong relationships \cite{preston2002empathy, morelli2017empathy, ong2015affective}. 
Indeed, these are some of the capabilities that affective computing research aims to achieve in AI agents. 
Although we are still far from having conversational affective agents that understand emotional narratives at the level of human listeners, there has been considerable progress in recent years, especially in using deep neural network models for naturalistic emotion recognition (see \cite{zeng2009survey, poria2017review} for reviews). For example, Convolutional Neural Networks (CNNs) have shown great success at recognizing emotions from facial expressions \cite{levi2015emotion} and from natural language text \cite{dos2014deep}.

Learning how to predict emotions continuously over time---i.e., being able to handle \emph{time-series} data---is essential for understanding naturalistic emotions \cite{nicolaou2011continuous, ong2019SEND}. 
The most popular class of deep learning time-series models are Recurrent Neural Networks (RNNs) \cite{williams1989learning} and Long Short-Term Memory (LSTM) networks
\cite{hochreiter1997long}: These networks incorporate neuronal units with a directed \emph{recurrent} connection to subsequent units, which enable them to model sequences over time. Many researchers have successfully applied RNNs \cite{kahou2015recurrent, khorrami2016deep} and LSTMs \cite{wollmer2008abandoning, eyben2010line, wollmer2013lstm, tan2019Multimodal} to recognize emotions from video.

Recently-proposed \emph{attention} mechanisms in deep neural network models have also shown great promise in learning to predict complex time-series data \cite{luong2015effective, bahdanau2015neural}. 
When people listen to a story, or even a string of words, not every word is equally important: People tend to pay attention to certain key words or phrases that carry relatively more information. Attention mechanisms in deep neural network models attempt to capture the intuition behind such behavior by trying to learn the relative importance of words within a given time-window. In vanilla implementations, attention layers in a deep model learn sets of weights over their inputs which are then used to upweight certain parts of the input over others.
These attention mechanisms have experienced much success, perhaps most exemplified by the recently-proposed Transformer model \cite{vaswani2017attention}, which currently represents the state-of-the-art on natural language sequence-prediction tasks. What is perhaps most interesting about the Transformer model is that it is entirely attention-based; It contains no recurrency (i.e., directed connections between hidden states at consecutive time-steps), which has been the de-facto ``standard" in sequence-prediction models to learn time-dependent information.
Since its introduction, Transformer-based models have been used in various NLP tasks, including text comprehension \cite{yu2018qanet}, Question-and-Answering \cite{devlin2018bert} machine translation \cite{gu2017non}, and language modelling \cite{radford2018language}. Notably, however, self-attention mechanisms like in the Transformer have not yet been applied to emotion recognition, and the very recent success of other types of attention applied to emotion recognition (e.g., in RNNs \cite{mirsamadi2017automatic} and LSTMs \cite{tan2019Multimodal}, or the Memory Fusion Network of \cite{zadeh2018memory}) suggest that these may be fruitful approaches that should be further investigated.

In this paper, we explore how attention mechanisms can be successfully applied to model emotion recognition from rich narrative videos. We propose and test two models (and several baselines) based on state-of-the-art attention models: the Transformer \cite{vaswani2017attention} for encoding input and the Memory Fusion Network \cite{zadeh2018memory} for multimodal fusion. We find that the Transformer is excellent at encoding information, but we find significant benefits in performance from adding recurrency. 
Using a recently-introduced dataset\cite{ong2019SEND} of individuals describing emotional life events, 
we find that deep neural networks with attention can achieve impressive results at recognizing the emotional valence, almost reaching human-level performance on some combinations of modalities, and we end by discussing the implications for future deep learning models.

\section{Model}

\begin{figure*}[!bt]
\centering
\includegraphics[width=1\textwidth]{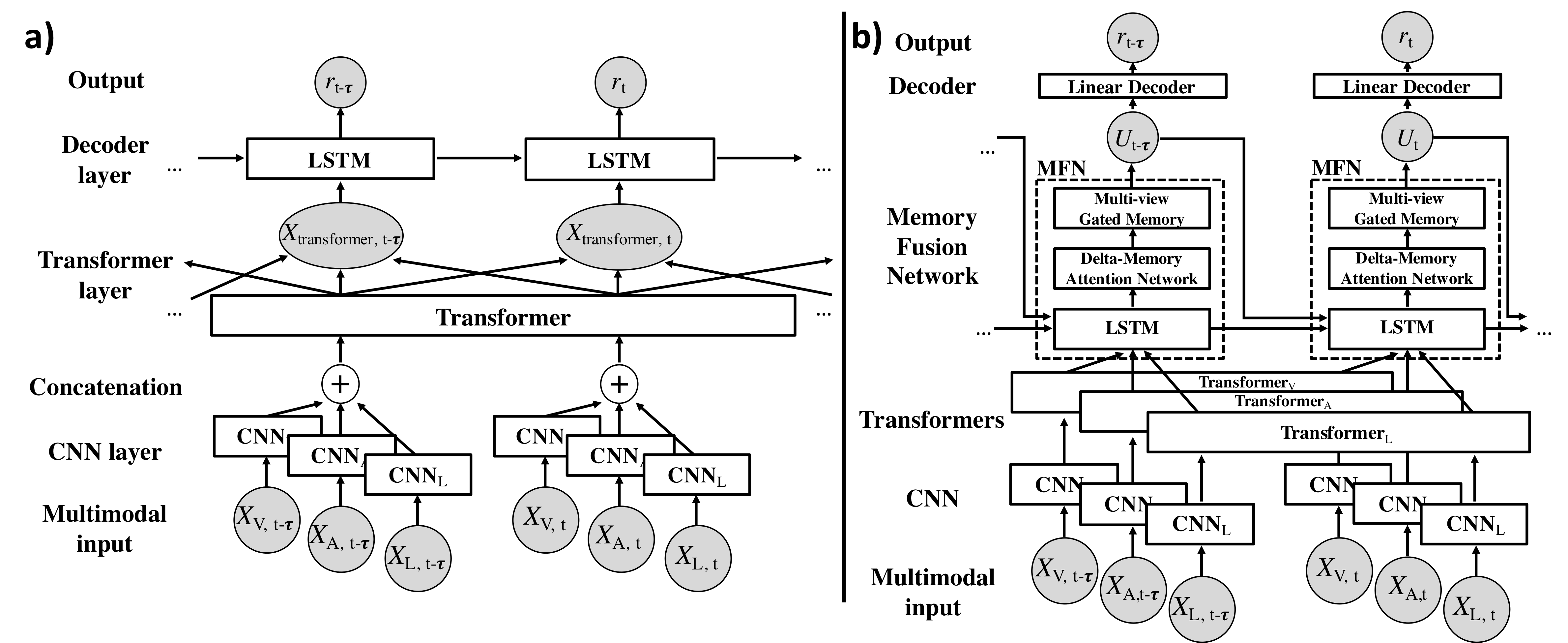}
\caption{Diagramatic overview of our two modelling approaches. (a) Simple Fusion Transformer. (b) Memory Fusion Transformer.}
\label{fig:Models}
\end{figure*}

In this work, in addition to the de-facto approach of \textit{recurrency} as implemented in an LSTM, we consider two ideas: \textit{self-attention} (as implemented in a Transformer), and \textit{cross-modality attention} (as implemented in a Memory Fusion Network, which also contains an LSTM). We find that combining recurrency and self-attention (in our \textbf{Simple Fusion Transformer, SFT}; Fig. \ref{fig:Models}a), and combining self-attention and cross-modality attention (in our \textbf{Memory Fusion Transformer, MFT}; Fig. \ref{fig:Models}b) performs extremely well; We further support this result by comparing the SFT and MFT to their individual components, implemented in three baseline models: \textbf{B1-LSTM}, \textbf{B2-Trans}, \textbf{B3-MFN}. 

As an overview, the SFT (Fig. \ref{fig:Models}a)
contains CNN, Transformer, and LSTM layers. The model uses CNN layers to processes the feature inputs from each modality to produce window embeddings. These embeddings from all modalities are then concatenated together using a linear layer with a tanh activation function. The fused window embeddings then enter a Transformer layer, and subsequently a LSTM decoder layer, which predicts a valence rating per window.

The MFT uses a Memory Fusion Network (MFN) to fuse multiple modalities. We train separate CNN-Transformer encoders for each modality; their outputs are then fed into a MFN that learns attention across modalities and time. Finally, we apply a linear decoder to produce one rating per window.

\subsection{Input Features}

The stimuli from the dataset we used (see Section \ref{sec:Dataset}) consist of multimodal videos of emotional narratives, along three modalities: \textbf{V}isual, \textbf{A}coustic, and \textbf{L}inguistic. 
For the \textbf{V}isual channel, we took frames every 0.1 second, and used openCV (v2.0.0) to crop and resize the face in each frame to 224$\times$224 px. We then fed these images into a pretrained VGG16 network \cite{simonyan2014very} and extracted 1000 features per frame from the final fully-connected linear layer. 
For the \textbf{A}coustic channel, we extracted 88 features as in the extended GeMAPS (eGeMAPS) \cite{eyben2016geneva}, for every second using openSMILE v2.3.0 \cite{eyben2013opensmile}. 
Finally, for the \textbf{L}inguistic features, we commissioned professional transcripts for all videos, then used forced alignment\footnote{https://github.com/ucbvislab/p2fa-vislab} to assign timestamps for each words, aligning the transcripts with the other channels. We then used 300-dimensional GloVe word embeddings \cite{pennington2014glove} as a representation for each word.

Multimodal time-series data are often sampled at different rates for each modality, and hence require synchronization \cite{gunes2005affect, snoek2005early}. 
One technique, is to undersample or oversample all modalities to the same sampling frequency. This allows multimodal inputs to be concatenated into a single vector at a given time-window \cite{chao2015long, chen2015multi, brady2016multi}. We adopt a similar approach, but rather than averaging samples over each time window to perform undersampling, we use one-dimensional CNNs to embed information over the samples in each time window. 


\subsection{Convolutional Neural Networks for embedding}
We use CNNs to produce window-level embedding vectors for each modality. Let $\textbf{v}_{\text{V}}$, $\textbf{v}_{\text{A}}$ and $\textbf{v}_{\text{L}}$ denote raw feature vectors for visual, acoustic, and linguistic inputs respectively. For ease of discussion, we consider the embedding of modality $m$, for $m \in [\text{V}, \text{A}, \text{L}]$, where each modality $m$ is sampled at different rates. Each vector $\textbf{v}_m$ is associated with a timestamp $t_{\textbf{v}_m}$.
Next we define a $\tau_m$-second-wide time-window, starting at a given time-point $t$, where we used $\tau_V\!=\!1$, $\tau_A\!=\!1$, and $\tau_L\!=\!5$. We stack the $n_m$ raw feature vectors $\textbf{v}_m$ that fall within the time-window (i.e. $t \leq t_{\textbf{v}_m} < t+\tau_m$), to create an input matrix $\textbf{X}_{m,t} \in \mathbb{R}^{|\textbf{v}_{m}|\times n^{\max{}}_m}$. Here, $n^{\max{}}_m$ is the maximum number of feature vectors that fall within any time-window for modality $m$ across the dataset.\footnote{When $n_m < n^{\max{}}_m$, as is often the case for the linguistic modality, we pad the remaining values by repeating the last feature vector.} We next feed $\textbf{X}_{m,t}$ through a one-dimensional CNN with kernel size $k\!=\!2$, and then perform max pooling across the time dimension:
\begin{align}
    \textbf{C}_{m,t} &= \text{Conv1D}(\textbf{X}_{m,t}) \quad &\in \mathbb{R}^{d_m \times ({n_m^{\max}}-k+1)} \\
    \textbf{X}_{\text{conv},m,t} &= \text{MaxPool}(\textbf{C}_{m,t}) \quad &\in \mathbb{R}^{d_m} 
\end{align}
Here we specify the output embedding dimensions, $d_{m}$, for visual, acoustic, and linguistic inputs to be 256, 256, and 300 respectively. Next, we synchronize the linguistic modality with the others by oversampling its window-level embeddings (i.e., repeating each 5-second window embedding five times), resulting in a common window size $\tau\!=\!1$s. We then apply a modified\footnote{We removed the ReLU operation from the original \emph{highway network}.} \emph{highway network} \cite{srivastava2015highway} which uses gating units to control information propagation through deep networks.
\begin{align}
    \textbf{X}_{\text{proj},m,t} &= \textbf{W}_{\text{proj},m}\textbf{X}_{\text{conv},m,t}+\textbf{b}_{\text{proj},m}  \\
    \textbf{X}_{\text{gate},m,t} &= \text{Softmax} \left( \textbf{W}_{\text{gate},m}\textbf{X}_{\text{conv},m,t}+\textbf{b}_{\text{gate},m} \right) 
\end{align}
with weight matrices $\textbf{W}_{\text{proj},m}$, $\textbf{W}_{\text{gate},m} \in \mathbb{R}^{d_m\times d_{m}}$ and bias vectors $\textbf{b}_{\text{proj},m}$, $\textbf{b}_{\text{gate},m} \in \mathbb{R}^{d_m}$. Finally, we obtain the output embedding of the CNN, $\textbf{X}_{\text{embed},m,t} \in \mathbb{R}^{d_m}$ by using a linear combination of the projection with a skip-connection:
\begin{align}
    \textbf{X}_{\text{embed},m,t} &= \textbf{X}_{\text{gate},m,t}\odot \textbf{X}_{\text{proj},m,t} \nonumber \\ &\quad + \left( 1-\textbf{X}_{\text{gate},m,t} \right)\odot \textbf{X}_{\text{conv},m,t}
\end{align}
where $\odot$ denotes element-wise multiplication. We trained our CNNs with dropout of probability 0.3.

\subsection{Transformer}

The Transformer \cite{vaswani2017attention} is a state-of-the-art neural network architecture for NLP tasks like machine translation \cite{devlin2018bert}. It uses a ``self-attention" mechanism to calculate an attention score for each token in a sequence. This allows the activation for a particular token within a network layer to depend upon the activations from all other tokens within that layer, i.e., a within-layer attention.

We use a Transformer as the encoder for our neural networks. After the CNNs, we have an embedding $\textbf{X}_{\text{embed}, m, t}$ where $m \in [\text{V}, \text{A}, \text{L}]$, for a window at time $t$. 
In our SFT model (Fig. \ref{fig:Models}a), we concatenate the embeddings from all $m$ modalities, $\textbf{X}_{\text{embed}, m, t}$,  into $\textbf{X}_{\text{fused},t} \in \mathbb{R}^{\sum_{m}(d_m)}$. The sequence \{$\ldots; \textbf{X}_{\text{fused},t};  \textbf{X}_{\text{fused},t+\tau}; \ldots$\} is input into the Transformer, which learns attention weights on each ``token" (time window).

\begin{figure}[!bt]
\centering
\includegraphics[width=1.00\columnwidth]{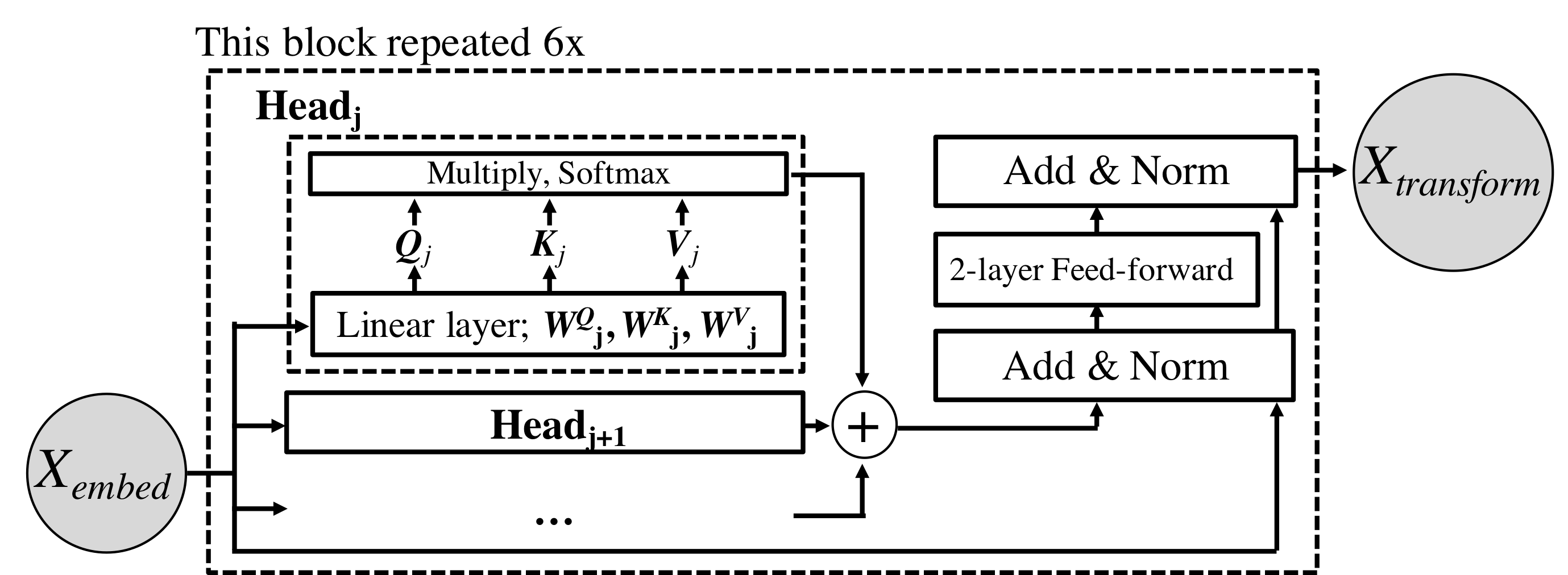}
\caption{Schematic of the basic Transformer architecture \cite{vaswani2017attention} we employed.}
\label{fig:Transformer}
\end{figure}

The basic building block of a Transformer is a \emph{multi-head} self-attention layer, followed by an element-wise feed forward layer. In the original paper, six of such blocks are stacked sequentially; Here, we also use six repeated blocks (Fig. \ref{fig:Transformer}). For an input \textbf{\textit{X}}, a \emph{single}-headed self-attention layer learns a set of weights \textbf{\textit{W}}$^\textbf{\textit{Q}}$, \textbf{\textit{W}}$^\textbf{\textit{K}}$ and \textbf{\textit{W}}$^\textbf{\textit{V}}$ that produces vector \emph{Queries}, \emph{Keys}, and \emph{Values}: In practice, these vectors are processed in parallel, so we have matrices \textbf{\textit{Q}}, \textbf{\textit{K}} and \textbf{\textit{V}}: 
\begin{align}
    \textbf{\textit{Q}} = \textbf{\textit{W}}^\textbf{\textit{Q}} \textbf{\textit{X}} \enskip;\enskip
    \textbf{\textit{K}} = \textbf{\textit{W}}^\textbf{\textit{K}} \textbf{\textit{X}} \enskip;\enskip
    \textbf{\textit{V}} = \textbf{\textit{W}}^\textbf{\textit{V}} \textbf{\textit{X}}
\end{align}
For a particular input token $j$, we consider the ``attention" that other input tokens $k$ bring to $j$ by multiplying the Query vector associated with $j$, $\textbf{\textit{Q}}_j$ with the Key vector associated with token $k$, $\textbf{\textit{K}}_k$. We process all of the tokens at once via the matrix multiplication $\textbf{\textit{Q}}\textbf{\textit{K}}^{T}$, scale by the reciprocal-square-root of the dimension of the query/key dimension $1/\sqrt{d_{k}}$ to keep the magnitudes of the product small, and take the Softmax. This term is then multiplied by the learnt Values $\textbf{\textit{V}}$:
%
%
\begin{align}
    \text{Attention}_{\textbf{\textit{W}}^\textbf{\textit{Q}}, \textbf{\textit{W}}^\textbf{\textit{K}}, \textbf{\textit{W}}^\textbf{\textit{V}}}\left( \textbf{\textit{X}} \right) &= \left( \text{Softmax}\left(
    \frac{\textbf{\textit{Q}}\textbf{\textit{K}}^{T}}{\sqrt{d_{k}}} \right)
    \right)
    \textbf{\textit{V}}
\end{align}

In the Transformer architecture, we employ multiple attention heads, resulting in \emph{Multi-head Attention}. Here we learn $h=8$ parallel attention heads, by initializing $h$ sets of single-headed attention parameters. For each head, we specified the dimension of the keys, queries and values to be 
$\sum_{m}(d_m) / h$.
The values after passing through each attention head are concatenated and multiplied by an Output Weight matrix \textbf{\textit{W}}$^\textbf{\textit{O}}$ to produce the output of the multiheaded attention:
\begin{align}
    \text{MultiHead}(\textbf{\textit{X}}) &= \left( \text{Concat}(\text{Head}_{1},...,\text{Head}_{h}) \right) \textbf{W}^{O}\\
    \text{where Head}_i &\equiv 
    \text{Attention}_{ \textbf{\textit{W}}^{\textbf{\textit{Q}}_i}, \textbf{\textit{W}}^{\textbf{\textit{K}}_i}, \textbf{\textit{W}}^{\textbf{\textit{V}}_i}}\left( \textbf{\textit{X}} \right)
\end{align}
%
%
Finally, we add a fully connected feed-forward network with residual connections and layer normalization. Specifically, we have two linear projections with a single ReLU activation in between, i.e., a two-layer neural network $f_T$:
\begin{align}
    f_T(\textbf{\textit{X}}) &= \textbf{W}_2 \left[ \text{ReLU}(\textbf{W}_{1} \textbf{\textit{X}} + \textbf{b}_{1}) \right] + \textbf{b}_2
\end{align}
We added dropout of p=0.1 before and after the feed-forward layers, and between Transformer blocks, as in \cite{vaswani2017attention}. Thus, to summarize, the Transformer layer in Fig. \ref{fig:Models}a and \ref{fig:Models}b consists of 6 stacked blocks where the output of one block is fed into the successive block: Each block contains one Multi-head Attention and one feed-forward network (Fig. \ref{fig:Transformer}). The output is a sequence across time: \{$\ldots; \textbf{X}_{\text{transformer},t}; \ldots$\}.

\subsection{Long Short-Term Memory Networks}
As mentioned, many researchers have used LSTMs \cite{hochreiter1997long} to predict emotions over time
\cite{wollmer2008abandoning, eyben2010line, wollmer2013lstm, tan2019Multimodal}. In our SFT (Fig. \ref{fig:Models}a), we use an LSTM to decode the output of the Transformer, $\textbf{X}_{\text{transformer},t}$, to produce a rating, $\hat{r_t}$, for a time-window at time $t$. The LSTM takes as input both the output of the Transformer at that time-point, as well as its own hidden state at the previous time-point, $t-\tau$. Finally, we apply a fully-connected linear layer to predict a rating for the time-window.
\begin{align}
    h_{t} &= \text{LSTM}(h_{t-\tau}, \textbf{X}_{\text{transformer}, t}) \label{eqn:LSTM} \\
    \hat{r}_{t} &= \textbf{W}_{\text{decoder}}h_{t} + \textbf{b}_{\text{decoder}} 
\end{align}
\subsection{Memory Fusion Network}
Going beyond Simple Fusion (simply concatenating different modalities), we adapted the \emph{Memory Fusion Network} \cite{zadeh2018memory}, which has been applied to predict emotion-relevant values---sentiment, valence/arousal, as well as personality traits---from multimodal inputs. Our \textbf{Memory Fusion Transformer} (Fig. \ref{fig:Models}b) combines elements of LSTMs and attention, which were already present in our Simple Fusion Transformer architecture, but in a manner that learns \emph{cross-modality} attention. Specifically, the Transformer learns attention weights on time-windows across all time, but not across modalities within a time-window, while, as we shall see, the MFN learns attention across different modalities within a 2-time-window segment.

In our MFT, we trained separate Transformer encoders for each modality, to produce separate $\textbf{X}_{\text{transformer}, m}\in \mathbb{R}^{d_{m}}$ where $m \in [\text{A}, \text{L}, \text{V}]$, which are then input into the MFN. The first layer of the MFN consists of a system of LSTMs, one for each modality. Similar to Eqn. \ref{eqn:LSTM}, we have:
\begin{align}
    h_{m,t}, c_{m,t} &= \text{LSTM}_{m}(h_{m,t-\tau}, \textbf{X}_{\text{transformer}, m, t}) \label{eqn:MFN-LSTM}
\end{align}
where in addition to the hidden state for modality $m$ and time $t$, $h_{m,t}$, we also store the corresponding \emph{memory cell state} $c_{m,t}$ of the LSTM. 

The goal of the next layer in the MFN, the Delta-Memory Attention Network (DMAN), is to learn attention weights on the LSTM cell states across two consecutive time-windows and all modalities. For a given time window $t$, the memory cell states at that window $c_{m,t}$ and the preceding window $c_{m,t-\tau}$ are concatenated and input into a neural network $f_A$ to learn attention weights. Let us define $C_t$ to be the concatenation of $c_{m,t}$ and $c_{m,t-\tau}$ for all $m$; $A_t$ the attention weights learnt by $f_A$; and $D_t$ the output of the DMAN:
\begin{align}
    C_t &\equiv \text{Concat}\left(c_{m,t}, c_{m,t-\tau}, \ldots \right) \quad m \in [A,L,V] \\
    A_t &= f_A(C_t) \enskip \qquad \qquad \qquad \qquad \text{attention weights} \\
    D_t &= A_t \odot C_t
\end{align}
For completeness, we note that $C_t, A_t, D_t \in \mathbb{R}^{2\sum_m d_m}$, or twice the sum of the modality-embedding dimensions, as they are across two time-windows. The output of the DMAN, $D_t$, is an attention-weighted memory cell state, which allows the DMAN to learn to ``attend" to certain parts of the memory cell states (by multiplying them with a higher weight) over others. The DMAN is trained with dropout of $p$=0.2.

The final layer in the MFN, the Multi-View Gated Memory (MGM), works in a similar manner to a vanilla LSTM, and propagates a multimodal ``memory state" $u_t$ over time. There are two gates $\gamma_1$ and $\gamma_2$ that respectively control how much of the previous state $u_{t-\tau}$ to retain, and how much to update with the output of the DMAN. These gates $\gamma_1$, $\gamma_2$, as well as the proposed update to the current MGM state $\hat{u}_t$, are learnt using multi-layer neural networks $f_{\gamma_1}, f_{\gamma_2}, f_{u}$:
\begin{align}
\gamma_{1,t} &= f_{\gamma_1}( D_t ) \qquad \qquad \qquad \qquad \text{retain gate} \\
\gamma_{2,t} &= f_{\gamma_2}( D_t ) \qquad \qquad \qquad \qquad \text{update gate} \\
\hat{u}_{t} &= f_{u}( D_t ) \enskip \qquad \qquad \qquad \qquad \text{proposed update}\\
u_t &= \gamma_{1,t}\odot u_{t-\tau} + \gamma_{2,t}\odot \text{tanh}(\hat{u}_t) \quad \text{update step}
\end{align}
Note that these gates are similar to an LSTM's forget and input gates, except that in a vanilla LSTM they are single-layer rather than multi-layer networks. The output of the MFN at each time-window $t$ is the concatenation of the MGM memory state $u_t$ and the LSTM hidden states $h_{m,t}$ (from Eqn. \ref{eqn:MFN-LSTM}). Recall that $h_{m,t}$ are still separate for each modality $m$; Ideally, after passing through the DMAN and the MGM layers, which apply attention across modalities and time as well as memory over time, the MGM memory state $u_t$ would contain fused multimodal information. This concatenated vector $U_t$ is then fed into a dropout layer ($p$=0.5) before a final linear decoder that produce a single rating:
\begin{align}
U_t &\equiv \text{Concat}(u_t, h_{m,t}, \ldots) \quad \quad m \in [A,L,V] \\
\hat{r}_t &= \textbf{W}_{\text{MGM}} ( U_t ) + \textbf{b}_{\text{MGM}}
\end{align}

\subsection{Baseline Models}

To compare the relative advantages of using the Transformer and the MFN in combination, we built three lesioned models to provide baseline comparisons. In our first baseline model, \textbf{B1-LSTM}, we removed the Transformer layer from our Simple Fusion Transformer (Fig. \ref{fig:Models}a): The input from each modality goes through a CNN, is concatenated, and fed into an LSTM.

In \textbf{B2-Trans}, we removed the LSTM layer from our Simple Fusion Transformer (Fig. \ref{fig:Models}a), replacing it with a simple linear decoder layer. Importantly, B2-Trans has \textit{no recurrent connections} between time-windows. 
In \textbf{B3-MFN}, we removed the Transformer layer from our Memory Fusion Transformer (Fig. \ref{fig:Models}b). The output from the CNN layer is fed directly into the Memory Fusion Network layer and a final linear decoder. 

\section{Stanford Emotional Narratives Dataset}
\label{sec:Dataset}

We previously introduced the Stanford Emotional Narratives Dataset (SEND) in \cite{ong2019SEND} and \cite{ong2017dissertation}, where we discuss the data collection procedure in greater detail. The \emph{SEND} comprises video recordings of participants (``targets") narrating emotional life events. These events were unscripted and varied in their content: Targets talked about positive events like winning a prize or going on vacation, to negative events like having a loved one pass away or experiencing a romantic breakup. This gives us a rich corpus that capture spontaneous variations in emotional content as well as emotional expression.

The \emph{SEND} consists of 193 video clips from 49 unique targets. On average, each clip lasted 2 mins 15 secs, for a total duration of 7 hrs and 15 mins. We created three partitions: a \textbf{Train set} (60\% of the dataset, 114 videos, 29 targets, 4 hrs 20 mins), a \textbf{Validation set} (20\%, 40 videos, 10 targets, 1 hr 29 mins) and a \textbf{Test set} (20\%, 39 videos, 10 targets, 1 hr 26 mins). Each target appeared in only one partition, to test the generalizability of our models to novel targets.

We further recruited a separate group of participants (``observers") to watch these clips and rate how they thought the target was feeling as they were speaking in the video. They made these annotations using a visual analog slider from ``Very Negative" [-1] to ``Very Positive" [1], sampled every 0.5s, giving us time-series ratings of emotional valence. We collected an average of 20 annotations per clip. To serve as the ``gold-standard" rating, we calculated the Evaluator Weighted Estimator (EWE \cite{grimm2007primitives}) of observers' ratings, which weights each observer $j$'s ratings $r^j$ by their (Pearson) correlation with the unweighted average $\overline{r}$:
\begin{align}
    r_{\text{EWE}} \equiv \frac{1}{\sum_j w^j} \sum_j w^j r^j \quad ; \quad
    w^j = \text{Corr}(r^j, \overline{r})
     \label{eqn:EWE}
\end{align}




To evaluate our models, we use the Concordance Correlation Coefficient (CCC \cite{lin1989concordance}), a commonly-used metric in affective computing \cite{valstar2016avec, ringeval2017avec}. The CCC for vectors $X$ and $Y$ is:
\begin{align}
    \text{CCC}(X,Y) &\equiv 
    \frac{2 \text{Corr}\left(X,Y\right) \sigma_X \sigma_Y}{\sigma_X^2 + \sigma_Y^2 + \left( \mu_X - \mu_Y \right)^2}
\end{align}
where $\text{Corr}\left(X,Y\right) \equiv \text{cov}(X,Y)/(\sigma_X \sigma_Y)$ is the Pearson correlation coefficient, and $\mu$ and $\sigma$ denotes the mean and standard deviation respectively. 

\subsection{Human Benchmark Results}
\label{sec:Modelling:HumanBenchmark}

Having multiple ratings per clip allowed us to calculate a benchmark performance on this task---how well an individual observer predicts the EWE of all \emph{other} observers. Let $\mathcal{K}$ denote the set of observers for video $k$, $r^j$ denote observer $j$'s ratings and $r_{\text{EWE}}^{\mathcal{J}_k \setminus j}$ denote the EWE of all other observers except $j$ (i.e., the remaining $\left( |\mathcal{J}_k|-1 \right)$ observers), then the mean human CCC on video $k$ is:
\begin{align}
    \overline{\text{CCC}}_k = \frac{1}{|\mathcal{J}_k|} \sum_{j \in \mathcal{J}_k} \text{CCC}\left( r^j, r_{\text{EWE}}^{\mathcal{J}_k \setminus j} \right) \label{eqn:humanCCC}
\end{align}
The mean and standard deviation of the observer CCC was $.53 \pm .13$ on the \textbf{Train set}, $.47 \pm .15$ on the \textbf{Validation set}, and $.50 \pm .12$ on the \textbf{Test set}.

\section{Results}

\begin{figure}[!bt]
\centering
\includegraphics[width=\columnwidth]{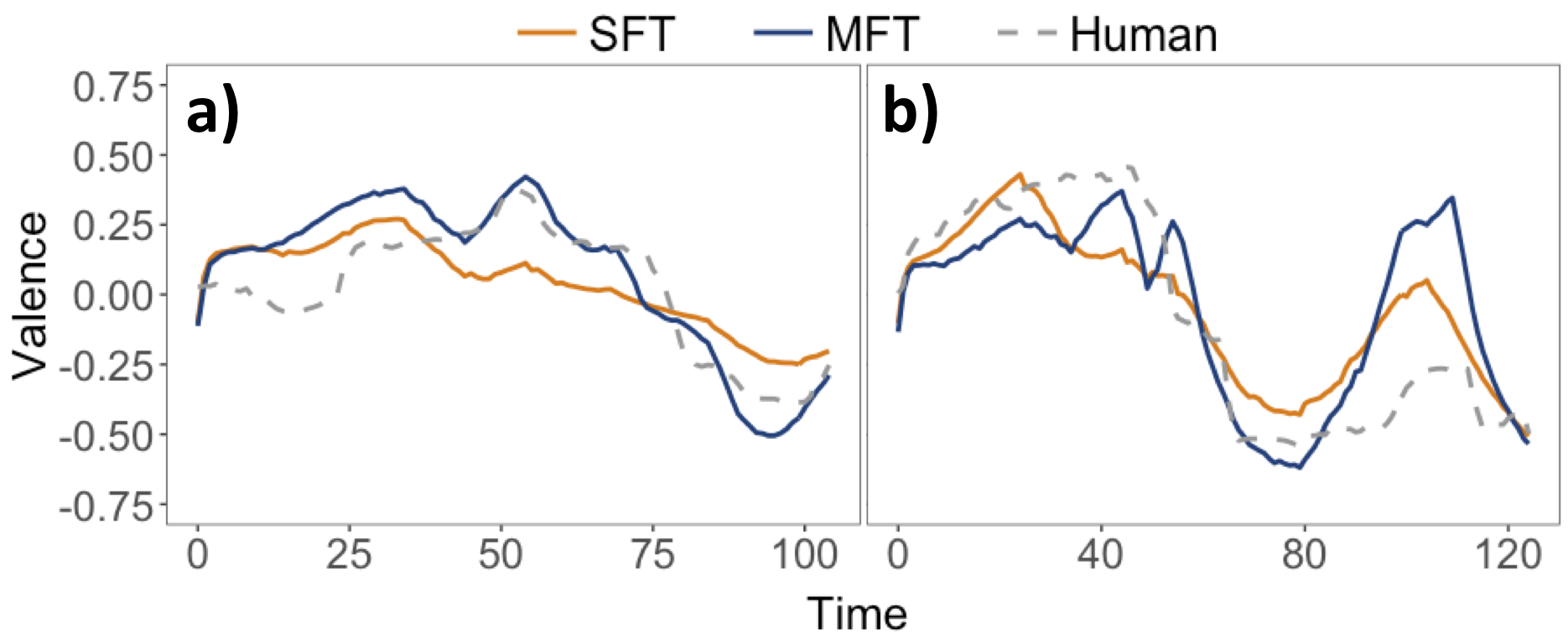}
\caption{Sample of the best-performing non-unimodal model predictions (SFT: Visual + Linguistic; MFT: Visual + Acoustic + Linguistic) compared with the mean-observer ratings. (a) is from the Validation set, (b) is from the Test set.}
\label{fig:Results}
\end{figure}

\begin{table}[!t]
    \scalebox{0.95}{
    \setlength\tabcolsep{2.5pt}
    \begin{tabular}{@{}llllllll@{}}
    \toprule
    \multirow{2}{*}{\textbf{Model}} & \multicolumn{7}{c}{\textbf{Modalities}} \\ 
    \cmidrule(l){2-8} 
     & \multicolumn{1}{c}{V} & \multicolumn{1}{c}{A} & \multicolumn{1}{c}{L} & \multicolumn{1}{c}{VA} & \multicolumn{1}{c}{AL} & \multicolumn{1}{c}{VL} & \multicolumn{1}{c}{VAL} \\ \midrule
    \multicolumn{8}{c}{Validation CCC (Std. Dev.)} \\ \midrule
    SFT & .12 (.23) & .15 (.34) & \textbf{.34 (.38)} & .15 (.27) & .08 (.19) & .32 (.28) & .12 (.27) \\
    MFT & -- & -- & -- & .06 (.19) & .36 (.31) & .40 (.31) & \textbf{.42 (.38)} \\
    B1-LSTM & .10 (.24) & .14 (.36) & .23 (.28) & .16 (.33) & .17 (.37) & .25 (.29) & .12 (.32) \\
    B2-Trans & .07 (.13) & .00 (.02) & .06 (.12) & .01 (.03) & .01 (.03) & .07 (.13) & .00 (.08) \\
    B3-MFN & -- & -- & -- & .22 (.32) & .37 (.30) & .33 (.28) & .34 (.31) \\
    Human & -- & -- & -- & -- & -- & -- & \textbf{.47 (.15)}  \\
    \midrule
    \multicolumn{8}{c}{Test CCC (Std. Dev.)} \\ \midrule
    SFT & .09 (.27) & .13 (.40) & .34 (.33) & .16 (.35) & .08 (.20) & \textbf{.35 (.31)} & .14 (.34) \\
    MFT & -- & -- & -- & .08 (.19) & .33 (.35) & .36 (.28) & \textbf{.44 (.31)} \\
    B1-LSTM & .05 (.17) & .09 (.33) & .21 (.22) & .06 (.31) & .17 (.34) & .17 (.21) & -.02 (.18) \\
    B2-Trans & .05 (.13) & .00 (.03) & .03 (.11) & .02 (.03) & .01 (.03) & .05 (.10) & .00 (.06) \\
    B3-MFN & -- & -- & -- & .09 (.33) & .33 (.30) & .31 (.30) & .28 (.30) \\
    Human & -- & -- & -- & -- & -- & -- & \textbf{.50 (.12)}  \\ \bottomrule
    \end{tabular}
    }
    \vspace{3pt}
    \caption{Summary of results. V: Visual, A: Acoustic, L: Linguistic. SFT: Simple Fusion Transformer (Fig. \ref{fig:Models}a), MFT: Memory Fusion Transformer (Fig. \ref{fig:Models}b). Human: See Sec. \ref{sec:Modelling:HumanBenchmark}. For SFT and MFT, we bold the best-performing combination. 
    }
    \label{tab:SummaryOfResults}
\end{table}

\subsection{Simple Fusion Transformer Results}
We summarize the results from all our models in Table \ref{tab:SummaryOfResults}, and plot sample predictions from the best-performing models in Fig. \ref{fig:Results}. Our Simple Fusion Transformer (Fig. \ref{fig:Models}a) performed the best in two modality combinations: when only using the Linguistic modality---mean CCC with standard deviation of .34 $\pm$ .38 on the Validation set and .34 $\pm$ .33 on the Test set---and when using a combination of Visual and Linguistics inputs---.32 $\pm$ .28 on Validation; .35 $\pm$ .31 on Test. 
Except for the Visual and Linguistics combination, the SFT does poorly on every other multimodal combination, in fact, significantly worse than Linguistics alone (paired $t$-tests comparing SFT\_L to other modalities; all $p$'s$<$.04). This is likely due to Simple Fusion---concatenating the multimodal embeddings---not adequately fusing information from multiple modalities.

Removing the Transformer from the SFT, as in our lesioned \textbf{B1-LSTM}, results in significantly worse performance on all modalities on the Test set (paired $t$-test, SFT$-$B1-LSTM across all modalities, $t(272)=3.43$, $p<.001$; for L modality only, $t(38)=2.30$, $p=0.03$). This suggests that the Transformer layer is essential for encoding emotional information, especially between different ``tokens" (time-windows) within a given video. 
%
%
On the other hand, the Transformer by itself, without a LSTM decoder, cannot capture all the information on this task either. \textbf{B2-Trans} consists of a Transformer with only a linear decoder, and this lesioned model does extremely poorly on every combination of modalities, significantly worse than the SFT ($t(272)=7.98$, $p<.001$) and even the B1-LSTM ($t(272)=5.03$, $p<.001$). The recurrent connections of the LSTM decoder layer is essential for propagating information across different time-windows. Thus, our results seem to suggest that it is the combination of \textbf{both} the Transformer and the LSTM layers that helps the SFT model achieve such high performance---though only on the Linguistics channel.


\subsection{Memory Fusion Transformer Results}

The inability of the SFT to better incorporate multiple modalities motivated us to explore more sophisticated fusion techniques by incorporating a Memory Fusion Network \cite{zadeh2018memory}. In the MFT, the embeddings produced by the Transformers (one for each modality) are fed into the Memory Fusion Network, which has as its first layer an LSTM. 
Our MFT performs excellently in any combination that includes the Linguistic channel, with the best-performing combination being trimodal---Visuals, Acoustics, and Linguistics---achieving a Validation CCC of .42 $\pm$ .38 and a Test CCC of .44 $\pm$ .31. 
The MFT significantly outperforms the SFT on every bi- and trimodal combinations on the Test Set (paired $t$-test: $t(155)$=$3.66$, $p<.001$).
The MFT's trimodal performance is also not significantly different from the human-level benchmark on the Test Set of .50 $\pm$ .12 ($t(38)$=$1.28$, $p$=$0.21$ n.s.). 

What if we only consider the MFN by itself? Removing the Transformer layer from the MFT, in \textbf{B3-MFN} (i.e., using only a CNN as an encoder into the MFN), results in significantly worse performance ($t(155)$=$2.07$, $p$=$0.04$) which is driven mainly by the trimodal case ($t(38)$=$2.83$, $p$=$.007$).
\textbf{B3-MFN} by itself does very well, not significantly different from the SFT's best bimodal combination ($t(38)$=$1.15$, $p$=$0.26$ n.s.). This suggests that the MFN alone can capture multimodal information well on this task, although its performance is improved by the addition of the Transformer.


\begin{figure}[!bt]
\centering
\includegraphics[width=\columnwidth]{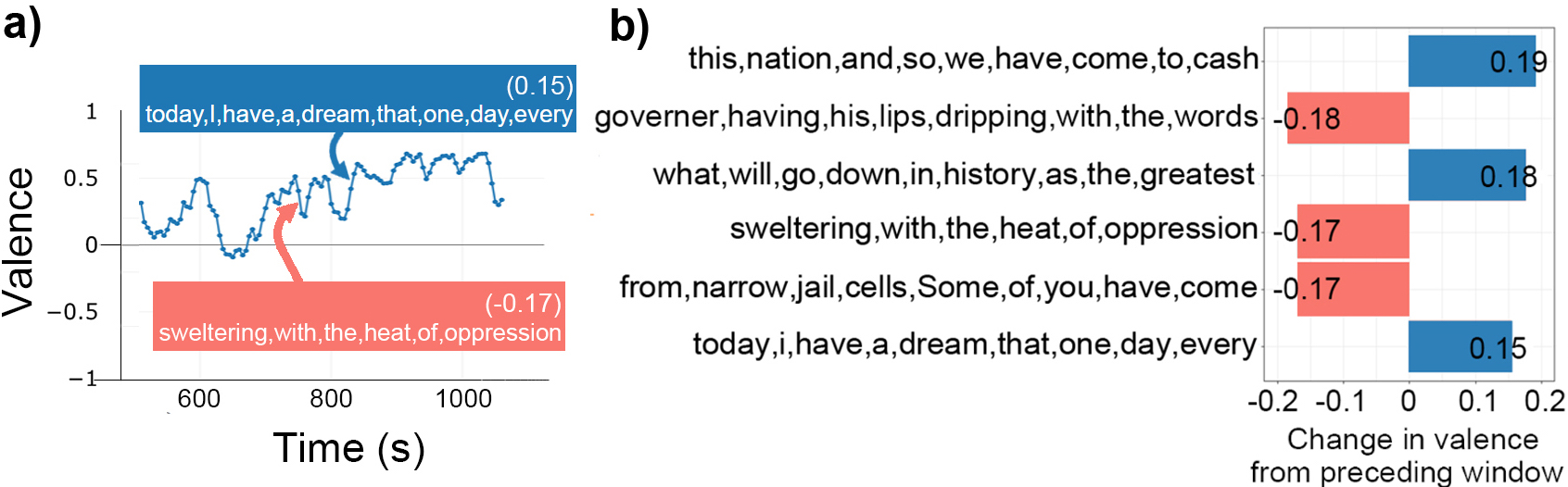}
\caption{Out-of-sample emotion valence prediction of the speech \textit{I Have A Dream} by the American baptist minister Martin Luther King Jr. with a Linguistics-only SFT. (a) Predictions of emotion valence of a segment of the speech, assuming a speaking rate of 0.1 seconds per character. We highlight two windows, along with their change from the previous window's valence. (b) Windows that the model predicted had the \emph{greatest change} in emotional valence from the previous window. We list the top 6.} 
\label{fig:dreamPlot}
\end{figure}

\section{Discussion}

It is challenging to recognize emotions in a natural setting like a conversation: There are many important signals in what people are saying, how they are saying it, and in nonverbal cues. Humans naturally pay attention to which cues across modalities matter more to one's emotion at any given instant, effectively and efficiently \emph{integrating} multiple modalities \cite{ong2015affective, ong2019computational, zaki2013cue}. Recently-proposed attention mechanisms in deep neural networks are inspired by such human attention, and attempt to learn the relative importance of input features, or hidden intermediate representations, in the networks \cite{luong2015effective, bahdanau2015neural}. In this work, we adapted and applied the latest in attention mechanisms to emotion understanding from naturalistic narratives. Our models combined two capabilities: being able to attend differentially to different inputs across time \cite{vaswani2017attention}, and across different modalities \cite{zadeh2018memory}. These attention mechanisms, coupled with recurrency (\`{a} la LSTM models), enabled our architectures to perform excellently at predicting emotional valence over time, in some instances coming very close to the human-level benchmark.

Our model also generalizes well out-of-sample. To demonstrate this, we chose a famous, emotionally-laden monologue: Martin Luther King Jr.'s \emph{I Have A Dream} speech. We took a Linguistic-only SFT model trained on our \emph{SEND}, and had it predict the emotional valence that Dr. King might have felt while giving his speech (Fig. \ref{fig:dreamPlot}a). Although we have no benchmark ratings, we can still visualize the valence predictions, as well as identify certain important time-windows that the model predicted had the greatest change in valence compared to the preceding time-window. Presumably, these are points that the model said (or a human might say), ``Here is an important window to pay attention to". And from a qualitative assessment, these windows do seem to be more emotionally charged. At present, we cannot directly visualize the ``attention" in the Transformer network (as it is applied over CNN embeddings, not input words), but we believe that efforts like this to probe what the model actually learns will be a fruitful area for future research. It may give insight into the inner workings of such deep models, while at the same time contributing towards building explainable affective computers.

Attention is a powerful idea in deep learning. At one level, if we think of emotion understanding as a signal processing problem, extracting signal from noise, then attention may allow one method to upweight certain parts of one modality, or even whole modalities over others, dynamically over time. Although attention in deep networks is not well understood---e.g., it is still unclear under what theoretical conditions attention is useful---and is likely very different from how human attention is actually implemented in the brain, these attention mechanisms have proven to be surprisingly effective in improving deep neural network performance. Aside from a few very recent papers 
\cite{zadeh2018memory, tan2019Multimodal, mirsamadi2017automatic}, there has not been much ``attention" paid to these attention mechanisms within affective computing. We hope that our results will help to demonstrate the efficacy of such approaches and to encourage more research in this area.


\section*{Acknowledgment}

We thank Isabella Kahhale and Alison Mattek for various assistance with the \emph{SEND}.

\bibliographystyle{IEEEtran}
\bibliography{biblio}

\end{document}